%% file: main.tex
\pdfoutput=1

\documentclass[11pt]{article}

\usepackage[]{acl}

\usepackage{times}
\usepackage{latexsym}

\usepackage[T1]{fontenc}

\usepackage{CJKutf8}
\usepackage[utf8]{inputenc}

\usepackage{microtype}

\usepackage{inconsolata}

\usepackage{soul}
\usepackage{subfigure}
\usepackage{bm}
\usepackage{amsmath}
\usepackage{amssymb}
\usepackage{tablefootnote}
\usepackage{multirow}
\usepackage{booktabs}
\usepackage{arydshln}
\usepackage{tikz}
\usetikzlibrary[arrows.meta, decorations.pathmorphing, backgrounds, positioning, fit, petri, graphs, math, external, calc, shapes, shapes.geometric, decorations.text, shadings]
\usepackage{tikz-dependency}
\usepackage{graphicx}
\usepackage{xcolor}
\usepackage{microtype}
\usepackage{mathtools}
\usepackage{adjustbox}
\usepackage{enumerate}
\usepackage{pgfplots}
\usepackage{fontawesome}
\usepackage{xpinyin}

\definecolor{brickred}{HTML}{b92622}
\definecolor{midnightblue}{HTML}{005c7f}
\definecolor{salmon}{HTML}{f1958d}
\definecolor{burntorange}{HTML}{f19249}
\definecolor{junglegreen}{HTML}{4dae9d}
\definecolor{forestgreen}{HTML}{499c5e}
\definecolor{pinegreen}{HTML}{3d8a75}
\definecolor{seagreen}{HTML}{6bc1a2}
\definecolor{limegreen}{HTML}{97c65a}
\definecolor{pink}{HTML}{E8A0BF}
\definecolor{purple}{HTML}{7149C6}

\title{CopyNE: Better Contextual ASR by Copying Named Entities 
}

\author{
    Shilin Zhou$^{1}$,
    Zhenghua Li$^{1}$\Thanks{$~$ Corresponding author},
    Yu Hong$^{1}$, \\
    \textbf{Min Zhang}$^{1}$,
    \textbf{Zhefeng Wang}$^{2}$,
    \textbf{Baoxing Huai}$^{2}$\\
    $^1$Institute of Artificial Intelligence, School of Computer Science and Technology, \\
    Soochow University, Suzhou, China \\
    $^2$Huawei Cloud, China \\
    \texttt{slzhou.cs@outlook.com; \{zhli13,hongy,minzhang\}@suda.edu.cn} \\
    \texttt{\{wangzhefeng,huaibaoxing\}@huawei.com}
}

\pgfplotsset{compat=1.17}
\begin{document}
\begin{CJK}{UTF8}{gkai}
\maketitle
\input{content/abstract}
\input{content/motivation}

\input{content/method}

\input{content/exp}

\input{content/related_work}

\input{content/conclusion}

\input{content/limit}

\input{content/ack}

\bibliography{anthology,custom}

\appendix

\vspace{+2ex}
\begin{center}
\Large \textbf{Appendices} 
\end{center}
\vspace{+1ex}

\section{Datasets} \label{app:data}
Aishell \cite{bu-2017-aishell} and ST-cmds\footnote{https://www.openslr.org/38/} are two widely used Chinese Mandarin datasets.
Aishell contains about 150 hours of speech. 
ST-cmds was built based on commonly used online chatting and user command speeches, which contains about 110 hours of speech.
For the English dataset, we utilize the portion of data that has been manually annotated with entities by \citet{YadavG0S20}, which comprises approximately 150 hours. The Eng dataset is built by extracting content from well-known English datasets, including Librispeech \cite{libri}, CommonVoice\footnote{https://en.wikipedia.org/wiki/Common\_Voice}, Tedlium \cite{rousseau-etal-2012-ted}, and Voxforge\footnote{https://en.wikipedia.org/wiki/VoxForge}.

Table \ref{table:dataset} shows the detailed statistics of the datasets used in our experiments.
``Sent'' means the number of instances. ``NE'' is the number of different named entities in the dataset and also the size of the contextual entity dictionary used during inference.
\input{tables/dataset}

\section{Parameter Settings} \label{app:para}
We use 80-dimensional log-mel acoustic features  with 25ms frame window and 10ms frame shift. 
The log-mel features are first fed into a 2D convolutional layer for downsampling and mapped to 256 dimensions before being inputted into the Audio Encoder. Both the audio encoder and decoder consist of 6 Transformer layers with 4 attention heads each. The NE Encoder is composed of three LSTM layers, with the input being a randomly initialized 256-dimensional embedding vector and the hidden size being 512. 
The relative weight $\lambda$ in Equation \ref{equ:final_loss} is set to 0.7.
The experiments are conducted on two NVIDIA A100 GPUs.

In addition, for the experiments on Whisper, we use Whisper-small model\footnote{https://huggingface.co/openai/whisper-small}, which includes 12 transformer layers in both its encoder and decoder, and is pre-trained on a total of 680,000 hours of multi-lingual and multi-task data.
We replace our audio encoder and decoder with the Whisper model and fine-tune the parameters of the entire model on our training set for a maximum of 20 epochs.
During fine-tuning, the initial learning rate for the Whisper model's parameters is set to 1e-5, while the learning rate for other parameters is set to 1e-3 with 10,000 warm-up steps.
During inference, we used beam search with a beam size of 5 and 10 for models with and without Whisper, respectively.

\input{figs/beta}
\section{Influence of Negative Entities in Training} \label{app:neg_ne_in_training}
During training, we construct an NE dictionary for each batch. To enhance the model's ability of copying correct entities, we sample additional negative examples.

Suppose the dictionary already contains $m$ entities, either real entities or pseudo sub-string entities.
We sample $\beta \cdot m$ entities as negative examples from the training set.
We utilize the parameter $\beta$ to control the number of negative examples.
Thus, we get the final dictionary for this batch which contains a total of  $(\beta+1) \cdot m$ entities.
As shown in Figure \ref{fig:beta}, adding 1 or 2 times the number of negative samples can reduce transcription errors.
Specifically, when $\beta=2$, the CER and NE-CER decreased by 0.42\% and 0.44\% compared to $\beta=0$. 
However, as $\beta$ continues to increase, the error rate started to rise.
We think that this is due to the presence of excessive noise. 
This causes the model to excessively focus on the negative samples, thus affecting its ability to accurately copy entities.
Therefore, we set $\beta$ to 2 during training.

\end{CJK}
\end{document}

%% file: content/abstract.tex
\begin{abstract}
End-to-end automatic speech recognition (ASR) systems have made significant progress in general scenarios. 
However, it remains challenging to transcribe contextual named entities (NEs) in the contextual ASR scenario.
Previous approaches have attempted to address this by utilizing the NE dictionary.
These approaches treat entities as individual tokens and generate them token-by-token, which may result in incomplete transcriptions of entities.
In this paper, we treat entities as indivisible wholes and introduce the idea of copying into ASR. 
We design a systematic mechanism called CopyNE, which can copy entities from the NE dictionary.
By copying all tokens of an entity at once, we can reduce errors during entity transcription, ensuring the completeness of the entity. 
Experiments demonstrate that CopyNE consistently improves the accuracy of transcribing entities compared to previous approaches.
Even when based on the strong Whisper, CopyNE still achieves notable improvements.

\end{abstract}

%% file: content/motivation.tex
\section{Introduction}
End-to-end automatic speech recognition (ASR) systems have achieved impressive performance in general scenarios \cite{chan-2016-las, rao-2017-exploring, gulati-2020-conformer, boulianne-2022-phoneme}.
However, in the contextual ASR scenario where speech often contains numerous contextual entities, it remains a challenge for ASR systems to get accurate transcriptions \cite{alon-2019-hard-example, jayanthi-2023-retrieve}.
For instance, when utilizing personal voice assistants like Siri or Alexa, it is common to encounter contextual entities such as personal names, place names, and organization names.
ASR models trained solely on speech-text data often struggle to transcribe these personalized entities due to their infrequent occurrence in the training set \cite{sathyendra-2022-contextual}.
Since contextual entities always cover a wealth of semantic information.
It is important to improve the accuracy of transcribing entities for downstream natural language processing tasks such as information retrieval and spoken language understanding \cite{ganesan-2021-nbest, wu-2022-towards-relation}.

\input{figs/intro-error}

Recently, researchers have started leveraging the information of textual modality as additional contextual knowledge to help contextual ASR.
The most typical approach, premised on the assumption that entities are already known before, use a contextual named entity (NE) dictionary as contextual knowledge \cite{chen-2019-joint_phone,jain-2020-contextual-rnnt,han-2021-cif,huber-2021-instant,char-level-concder-2023}.
Two representative approaches are ``contextual listen, attend and spell'' (CLAS) \cite{pundak-2018-clas} and contextual bias attention (CBA) \cite{zhang-2022-cba}.
CLAS employs the knowledge of the dictionary to aid the prediction of each token.
They use dictionary representation as extra inputs for token prediction in the decoder. 
The decoder attends to each entity, and the dictionary representation is an aggregated representation of all entities, weighted by the attention scores.
CBA extends CLAS and uses an extra training loss. 
The loss explicitly makes use of the attention scores and force the model attend to a proper entity in the dictionary if the token to be predicted is related with the entity.

Previous methods have achieved considerable improvements, especially in transcribing entities.
However, they all treat entities as individual tokens.
These models utilize contextual knowledge to aid in predicting independent tokens without considering the role of these tokens in constituting a complete entity.
In other words, a multi-token entity is broken into isolated tokens during decoding.
We argue that this is problematic. 
For instance, model may erroneously generate the subsequent tokens of an entity, despite correctly producing the preceding tokens.
As shown in Figure \ref{fig:intro-error}, when transcribing the speech ``他来自安徽铜陵'' (He comes from Anhui Tongling), an incorrect output of ``他来自安徽铜铃'' (He comes from Anhui copper bell) is obtained.
Despite the model's awareness of the location entity ``铜陵'' in the NE dictionary and its accurate prediction of the first token ``铜'', it mistakenly transcribes ``陵'' (ling) as ``铃'' (bell) during the token-level prediction process.
This occurs because the model predicts tokens independently, neglecting the integrity of the token span as a complete entity.
Furthermore, ``陵'' and ``铃'' share the same pronunciation ``\pinyin{ling2}'', with ``铃'' being a more frequently occurring token in the training data.
Consequently, the model tends to generate the wrong token ``铃''.

In this paper, we propose a new approach for contextual ASR called CopyNE. Unlike previous approaches, we view entities as indivisible wholes. 
To the best of our knowledge, we are the first to introduce the idea of copying into ASR.
We design a systematic and effective mechanism to copy entities from a dictionary. 
Specifically, CopyNE uses a copy loss that guides the model to copy the correct entity from the dictionary. 
During inference, our CopyNE has the flexibility to either predict a token from the token vocabulary or copy an entity from the NE dictionary at each decoding step. 
By copying multiple tokens simultaneously, we can alleviate errors within the entity, thus ensuring the token span as a complete entity.

Experiments on Chinese Aishell \cite{bu-2017-aishell}, ST-cmds\footnote{https://www.openslr.org/38/}, and English Eng \cite{YadavG0S20} datasets show that our CopyNE achieves significant improvements across all scenarios, particularly in the contextual ASR scenario. 
Compared to previous methods using dictionary, CopyNE achieves relative reductions in CER of 13.5\% and 20.8\% on Aishell and ST-cmds in the contextual scenario.
Notably, CopyNE shows more remarkable improvements when it comes to transcribing entities, with relative reductions of 55.4\% and 53.9\% in the NE-CER metric on Aishell and ST-cmds.
Moreover, when based on Whisper \cite{radford-2022-whisper} and evaluated in its domain of expertise, Eng dataset, CopyNE still achieves an impressive 6.4\% and 16.8\% relative reductions in WER and NE-WER.
We will release our codes, configurations, and models at \url{https://github.com/zsLin177/CopyNE}.

%% file: figs/intro-error.tex
\begin{figure}[!tb]
    \centering
    \scalebox{0.7}{
    \begin{tikzpicture}[
        connect/.style={
                rounded corners=4pt,
                very thick,
                draw=black!80
            },
        dashed connect/.style={
                rounded corners=4pt,
                densely dashed,
                draw=black!80
            },
        arrow/.style={
                arrows = {-Straight Barb[length=0.8mm]},
                shorten >= 2pt,
                shorten <= 1.5pt,
                very thick
            },
        inner arrow/.style={
                arrows = {-Straight Barb[length=0.4mm]},
                shorten >= 2pt,
                shorten <= 2pt,
                thin,
                draw=black!50
            },
        input/.style={
                rectangle,
                rounded corners=1mm,
                thin,
                dashed,
                draw=none,
                minimum width=3.5cm,
                minimum height=0.6cm,
            },
        share/.style={
                minimum height=0.5cm,
                  fill=orange!10,
                draw={rgb,255:red,225; green,209; blue,183},
                rounded corners=2mm,
            },
        chart/.style={
                circle,
                minimum size=3mm,
                draw=white,
                fill={rgb,255:red,189; green,215; blue,237},
                inner sep=0pt
        },
        chart1/.style={
                circle,
                minimum size=3mm,
                draw=white,
                fill={rgb,255:red,30; green,78; blue,120},
                inner sep=0pt
        },
        chart2/.style={
                circle,
                minimum size=3mm,
                draw=white,
                fill={rgb,255:red,205; green,223; blue,230},
                inner sep=0pt
        },
        chart3/.style={
                circle,
                minimum size=3mm,
                draw=white,
                fill={rgb,255:red,68; green,113; blue,199},
                inner sep=0pt
        },
        vilayer/.style={
                trapezium, 
                draw, 
                trapezium left angle=60, 
                trapezium right angle=120,
                rounded corners=1mm,
        },
        task4/.style={
                minimum height=0.5cm,
                draw=black,
                rounded corners=2mm,
                fill={rgb,255:red,240; green,249; blue,244},
            },
        task3/.style={
                minimum height=0.5cm,
                draw=white,
                rounded corners=2mm,
            },
        task2/.style={
                minimum height=0.5cm,
                fill={rgb,255:red,204; green,223; blue,230},
                draw={rgb,255:red,151; green,181; blue,191},
                rounded corners=2mm,
            },
        label/.style={
                inner sep=0.5mm,
                fill=white,
                minimum height=0.5cm,
            },
        task1/.style={
                minimum height=0.5cm,
                fill={rgb,255:red,253; green,210; blue,191},
                draw={rgb,255:red,233; green,133; blue,128},
                rounded corners=2mm
            },
        task5/.style={
                minimum height=0.5cm,
                fill={rgb,255:red,255; green,255; blue,255},
                draw={rgb,255:red,0; green,0; blue,0},
                rounded corners=2mm
            },
        task6/.style={
                minimum height=0.0cm,
                fill={rgb,255:red,255; green,255; blue,255},
                draw={rgb,255:red,0; green,0; blue,0},
                rounded corners=2mm
            },
        inner lstm/.style={
                fill=white,
                rectangle,
                rounded corners=1mm,
                semithick,
                draw=black!50,
                fill opacity=0.8
            },
        cell/.style={
                inner sep=2mm,
                rectangle,
                rounded corners=1mm,
                semithick,
                draw=black!50,
            },
        ocell/.style={
                solid,
                minimum height=0.5cm,
                rectangle,
                rounded corners=1mm,
                thick,
            },
        hatcell/.style={
                solid,
                minimum height=0.0cm,
                rectangle,
                rounded corners=0mm,
                thick,
            },
        dep arrow/.style={
        arrows = {-Latex[round,open,length=8pt,width=6pt]},
        shorten >= 2pt,
        shorten <= 1.5pt,
        thick
        }
        ]
        \centering
        \node[inner sep=0pt] (waveform) at (0,0)
            {\includegraphics[width=0.2\textwidth]{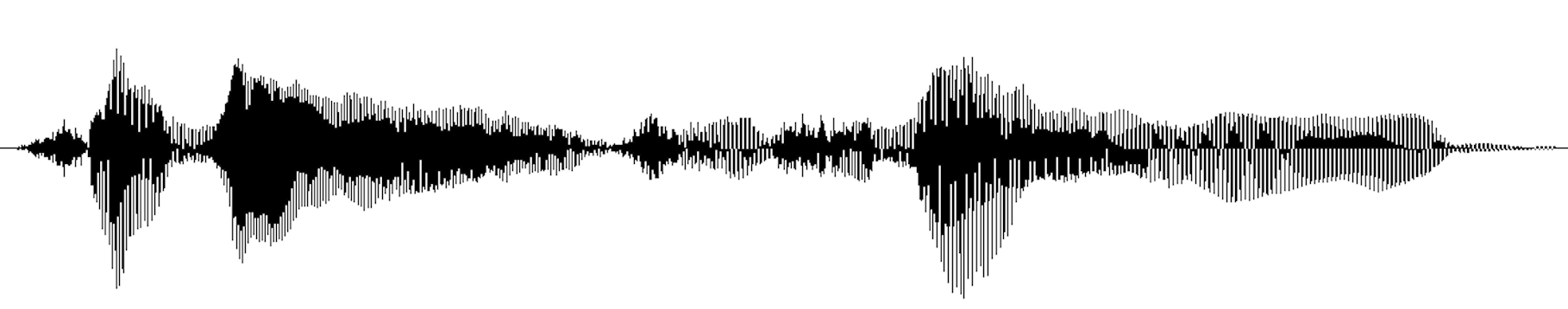}};
        \node[anchor=base] at ($(waveform.south) + (+0.cm, -0.5cm)$) (input_audio) {input audio};

        \node [task2, ocell, anchor=south, fill opacity=0.85, minimum height=1.cm, minimum width=2.5cm, align=center] (ASR) at ($(waveform.north) + (+0cm, 0.8cm)$) {ASR System};

        \draw [arrow, connect] ($(waveform.north) + (0cm, -0.0cm)$) -- ($(ASR.south) + (0cm, 0)$);

        \node[anchor=south] at ($(ASR.north) + (+0.4cm, +1.cm)$) (an) {\textcolor{black}{\large{安}}};
        \node[anchor=west] at ($(an.east) + (+0.2cm, 0.cm)$) (hui) {\textcolor{black}{\large{徽}}};
        \node[anchor=west] at ($(hui.east) + (+0.2cm, 0.cm)$) (tong) {\textcolor{black}{\large{铜}}};
        \node[anchor=west] at ($(tong.east) + (+0.2cm, 0.cm)$) (ling) {\textcolor{red}{\large{铃}}};
        \node[anchor=east] at ($(an.west) + (-0.2cm, 0.cm)$) (zi) {\textcolor{black}{\large{自}}};
        \node[anchor=east] at ($(zi.west) + (-0.4cm, 0.cm)$) (lai) {\textcolor{black}{\large{来}}};
        \node[anchor=east] at ($(lai.west) + (-0.3cm, 0.cm)$) (ta) {\textcolor{black}{\large{他}}};
        \draw [arrow, connect] ($(ASR.north) + (0.cm, -0.0cm)$) -- ($(an.south) + (-0.4cm, 0)$);
        \node[anchor=east] at ($(ta.west) + (-0.1cm, +0.cm)$) (output) {\textcolor{black}{\large{Output:}}}; 

        \node[anchor=south] at ($(an.north) + (+0.cm, +0.1cm)$) (py-an) {\textcolor{black}{\large{\pinyin{an1}}}};
        \node[anchor=south] at ($(zi.north) + (+0.cm, +0.1cm)$) (py-zi) {\textcolor{black}{\large{\pinyin{zi4}}}};
        \node[anchor=south] at ($(lai.north) + (+0.cm, +0.1cm)$) (py-lai) {\textcolor{black}{\large{\pinyin{lai2}}}};
        \node[anchor=south] at ($(ta.north) + (+0.cm, +0.1cm)$) (py-ta) {\textcolor{black}{\large{\pinyin{ta1}}}};
        \node[anchor=south] at ($(hui.north) + (+0.cm, +0.1cm)$) (py-hui) {\textcolor{black}{\large{\pinyin{hui1}}}};
        \node[anchor=south] at ($(ling.north) + (+0.cm, +0.02cm)$) (py-ling) {\textcolor{black}{\large{\pinyin{ling2}}}};
        \node[anchor=south] at ($(tong.north) + (+0.cm, +0.02cm)$) (py-tong) {\textcolor{black}{\large{\pinyin{tong2}}}};
         \node[anchor=east] at ($(py-ta.west) + (-0.1cm, +0.cm)$) (pinyin) {\textcolor{black}{\large{Pinyin:}}};

         \node[anchor=south] at ($(py-ta.north) + (+0.cm, +0.1cm)$) (g-ta) {\textcolor{black}{\large{他}}};
        \node[anchor=south] at ($(py-lai.north) + (+0.cm, +0.05cm)$) (g-lai) {\textcolor{black}{\large{来}}};
        \node[anchor=south] at ($(py-zi.north) + (+0.cm, +0.06cm)$) (g-zi) {\textcolor{black}{\large{自}}};
        \node[anchor=south] at ($(py-an.north) + (+0.cm, +0.06cm)$) (g-an) {\textcolor{black}{\large{安}}};
        \node[anchor=south] at ($(py-hui.north) + (+0.cm, +0.04cm)$) (g-hui) {\textcolor{black}{\large{徽}}};
        \node[anchor=south] at ($(py-tong.north) + (+0.cm, +0.03cm)$) (g-tong) {\textcolor{black}{\large{铜}}};
        \node[anchor=south] at ($(py-ling.north) + (+0.cm, +0.03cm)$) (g-ling) {\textcolor{black}{\large{陵}}};
        \node[anchor=east] at ($(g-ta.west) + (-0.1cm, +0.cm)$) (gold) {\textcolor{black}{\large{Gold:}}};

        \node[anchor=south] at ($(g-ta.north) + (+0.cm, +0.1cm)$) (he) {\textcolor{black}{\large{He}}};
        \node[anchor=south] at ($(g-lai.north) + (+0.cm, +0.06cm)$) (comes) {\textcolor{black}{\large{comes}}};
        \node[anchor=south] at ($(g-zi.north) + (+0.cm, +0.06cm)$) (from) {\textcolor{black}{\large{from}}};
        \node[anchor=south] at ($(g-an.north) + (+0.cm, +0.1cm)$) (An) {\textcolor{black}{\large{An}}};
        \node[anchor=south] at ($(g-hui.north) + (+0.cm, +0.1cm)$) (Hui) {\textcolor{black}{\large{hui}}};
        \node[anchor=south] at ($(g-tong.north) + (+0.cm, +0.02cm)$) (Tong) {\textcolor{black}{\large{Tong}}};
        \node[anchor=south] at ($(g-ling.north) + (+0.cm, +0.02cm)$) (Ling) {\textcolor{black}{\large{ling}}};
        \node[anchor=east] at ($(he.west) + (-0.1cm, +0.cm)$) (english) {\textcolor{black}{\large{English:}}};
        
        \node [task6, hatcell] [minimum width=1.8cm, minimum height=0.01cm, anchor=west, fill opacity=0.85, align=center] (z1) at ($(ASR.north east) + (1.cm, -0.2cm)$) {\footnotesize{铜 陵}};
        \node [task6, hatcell] [minimum width=1.8cm, minimum height=0.01cm, anchor=north, fill opacity=0.85, align=center] (z2) at ($(z1.south) + (0.cm, -0.01cm)$) {\footnotesize{\textcolor{white}{铜}}\large{...}\footnotesize{\textcolor{white}{铜}}};
        \node[anchor=base] at ($(z2.south) + (+0.cm, -0.5cm)$) (input_dictionary) {\small{NE dictionary}};
        \draw [arrow, connect] ($(z2.north west) + (0cm, -0.0cm)$) -- ($(ASR.east) + (0cm, +0.025cm)$);

        \draw [arrow, dashed connect] ($(tong.south) + (0cm, -0.0cm)$) -- ($(z1.north) + (-0.05cm, +0.cm)$);
        \draw [arrow, dashed connect, draw=red!80] ($(ling.south) + (0cm, -0.0cm)$) -- ($(z1.north) + (+0.05cm, +0.cm)$);

    \end{tikzpicture}}
    \caption{
    An example with homophonic errors. Pinyin is the Mandarin pronunciation of each token.
    The red text indicates the wrongly predicted token.
    }
    \label{fig:intro-error}
\end{figure}

%% file: content/method.tex
\section{The CTC-Transformer Model}

In this work, we build our proposed approach on the end-to-end CTC-Transformer model, since it is the most widely used and achieves competitive performance in the ASR field \cite{hori-2017-jointctc, kim-2017-joint, haoran-2020-online, omachi-2021-end, gong-2022-train}.
However, it is worth noting that our idea can be applied to other ASR approaches as well.

The CTC-Transformer is built upon the seq-to-seq Transformer \cite{vaswani-2017-attention}, with a connectionist temporal classification (CTC) layer added after the audio encoder. 
As shown in Figure \ref{fig:joint-ctc-transformer-model}, it takes a sequence of acoustic frames $\bm{X} = (\bm{x}_1, ..., \bm{x}_T)$ as input and generates the corresponding transcription text $\bm{y} = (y_1, ..., y_U)$ as output. The model consists of two main components: an encoder and a decoder. First, the encoder encodes the acoustic frames $\bm{X}$ into hidden states $\bm{H} = (\bm{h}_1, ..., \bm{h}_T)$. Then, the decoder predicts the target sequence $\bm{y}$ in an auto-regressive manner. At each decoding step $u$, the decoder predicts the next target token $y_{u+1}$ based on the encoder's output $\bm{H}$ and the previously predicted tokens $y_{\leq u} = (y_1, ..., y_{u})$. This process is expressed as follows:

\input{figs/joint-ctc-transformer}
\begin{equation}
\bm{H} = \mathrm{AudioEncoder}(\bm{X})
\end{equation}
\begin{equation}
\bm{d}_u = \mathrm{Decoder}(y_{\leq u}, \bm{H})
\end{equation}
\begin{equation}\label{equ:raw_p}
P(y_{u+1}|y_{\leq u}) = \mathrm{softmax}(\bm{W}\bm{d}_u+\bm{b})
\end{equation}
Here, $\bm{d}_u \in \mathbb{R}^{d}$ denotes the hidden state at step $u$, and $P(y_{u+1}|y_{\leq u})$ is the posterior distribution of predicting token $y_{u+1}$. $\bm{W} \in \mathbb{R}^{|\mathcal{V}| \times d}$ and $\bm{b} \in \mathbb{R}^{|\mathcal{V}|}$ are learned parameters, where $\mathcal{V}$ is the token vocabulary, and $|\mathcal{V}|$ is the size of the vocabulary.

The loss of Transformer, $\mathcal{L}_{trans}(\bm{y})$, comes from minimizing the negative log probability of $\bm{y}$.
\begin{equation}
    \mathcal{L}_{trans}(\bm{y}) = -\sum_{u=0}^{U-1}\log P(y_{u+1}|y_{\leq u})
\end{equation}

As commonly used in previous works, the CTC loss is also applied here.
CTC aligns each acoustic frame with a token from left to right. For a given target sequence $\bm{y}$, there may be multiple valid alignments. The CTC loss is derived from maximizing the sum of these valid alignments, and has been proved to be able to enhance the representational capacity of the audio encoder \cite{kim-2017-joint}.
Finally, the overall loss is the a weighted sum of the $\mathcal{L}_{trans}(\bm{y})$ and $\mathcal{L}_{ctc}(\bm{y})$, as follows:
\begin{equation}
    \mathcal{L}(\bm{y}) = \lambda \mathcal{L}_{trans}(\bm{y}) + (1-\lambda)\mathcal{L}_{ctc}(\bm{y})
\end{equation}
where $\lambda$ is a hyper-parameter that determines the relative weight of each loss term.

In inference, the model selects the most probable transcription using beam search as follows:
\begin{equation}\label{equ:base_decoding}
    \hat{\bm{y}} = \arg \max \limits_{\bm{y}} (\sum_{u}\log P(y_{u+1}|y_{\leq u}))
\end{equation}
Here, there are many ways to use scores in decoding, such as combining CTC scores and Transformer scores as in training, or using CTC-prefix beam search followed by re-scoring with Transformer to select the optimal result. To compare with most previous works, we use the simplest decoding strategy, as shown in Equation \ref{equ:base_decoding}.

\section{Our CopyNE Model}
This section describes our proposed CopyNE model. 
The basic idea is that the model incorporates a contextual NE dictionary 
as external knowledge and can choose to directly copy NEs from the dictionary. 
We design a systematic framework to implement the idea.
During training, a copy loss is designed to %
encourage the model to copy corresponding entities from the dictionary.
During inference, at each generation step, the model can either predict a single token from token vocabulary or directly copy a entity from the given dictionary.

\subsection{The Model Framework}

Figure \ref{fig:copyne-model} illustrates the framework of our CopyNE model, which shares the same encoder as the CTC-Transformer model, but with a distinct decoder.
In the decoder, we introduce an extra NE encoder that takes the NE dictionary as input and encodes it into NE representations.
Then, we use a dot-product attention module to compute copy probabilities based on the obtained NE representations, which are then aggregated to form the overall dictionary (Dict) representation. The decoder can not only utilize copy probabilities to select entities for copying but also leverage the Dict representation to aid in predicting the next token.

\paragraph{NE Representation.}
We denote an NE dictionary as $E=(e_0, e_1, ..., e_N)$. We use $e_0=\varnothing$ as a pseudo entity to handle the case where the text to be transcribed has no relation to any entity and the model should not copy any entity at current step. 

For each entity $e_i$, we apply a multi-layer LSTM as the NE encoder to encode the token sequence and use the last hidden state of the NE encoder as the entity representation. It is a popular practice in previous {contextual ASR} works \cite{pundak-2018-clas, zhang-2022-cba}.
\begin{equation} \label{equ:lstm}
    \bm{z}_i = \mathrm{LSTM}(e_i)
\end{equation}
After that, we get entity representations $\bm{Z}=(\bm{z}_{0}, \bm{z}_{1}, ..., \bm{z}_{N})$, where $\bm{Z} \in \mathbb{R}^{N \times d}$.

\paragraph{Copy Probability.}
Once the NE representations are obtained, the copy 
probability is computed by a dot-product attention mechanism.
It is used to determine which entity to copy.
First, we compute the attention score $a_{u}^{e_{i}}$ for entity $e_{i}$ at step $u$ as follows:
\begin{equation} \label{equ:att_score}
    a_{u}^{e_{i}} = \cfrac{(\bm{W}_{q}\bm{d}_{u})^{\top} (\bm{W}_{k}\bm{z}_{i})}{\sqrt{d}}
\end{equation}
where $\bm{W_{q}}, \bm{W_{k}} \in \mathbb{R}^{d_a \times d}$ are two learned parameters. $d_a$ denotes the dimension of the attention.
After that we obtain the attention probability $P_{c}(e_{i}|y_{\leq u})$ for entity $e_{i}$ by softmax.
\begin{equation} \label{equ:att_prob}
    P_{c}(e_{i}|y_{\leq u}) = \cfrac{\exp(a_{u}^{e_{i}})}{\sum_{e_{j} \in E} \exp(a_{u}^{e_{j}})}
\end{equation}
Here, $P_{c}(e_{i}|y_{\leq u})$ not only represents the attention probability of $e_{i}$ but also naturally serves as the copy probability for the entity.
During inference, we use the copy probabilities to select the entities for copying.

\paragraph{Dict Representation.}
With copy (attention) probabilities, we can obtain the Dict representation $\bm{r}_{u}$ at decoding step $u$. It is used to help the prediction of subsequent tokens.
Specifically, $\bm{r}_{u} \in \mathbb{R}^{d}$ is computed by weighted summing the entity representations with the copy (attention) probabilities.
\begin{equation} \label{equ:dic_repr}
    \bm{r}_{u} = \sum_{e_{i} \in E} P_{c}(e_{i}|y_{\leq u}) \bm{z}_{i}
\end{equation}

\paragraph{Dict-enhanced Prediction.}
Finally, we get the overall Dict representation and copy probabilities.
Following \citet{pundak-2018-clas}, the Dict representation is applied to help the generation of the next token.
So Equation \ref{equ:raw_p} is extended as follows:
\begin{equation} \label{equ:new_p}
    P(y_{u+1}|y_{\leq u}, E) = \mathrm{softmax}(\bm{W}\left[\bm{d}_{u}, \bm{r}_{u}\right]+\bm{b})
\end{equation}

\subsection{Training}
During training, to guide the model in selecting correct entities from the NE dictionary for copying, 
we introduce an additional copy loss $\mathcal{L}_{copy}$.
First, based on the ground truth transcription $\bm{y}$ and the NE dictionary, we construct a copy target $\sigma_{u+1}$ for each decoding step $u$, telling the model whether to copy an entity from the dictionary or not, and which one to copy.
Then we compute the copy loss $\mathcal{L}_{copy}$ according to the copy target $\sigma_{u+1}$ and the copy probability $P_{c}(\sigma_{u+1}|y_{\leq u})$.

\paragraph{The Computation of Copy Loss.}
Provided that we have an NE dictionary $E^{b}$, we construct a copy target, denoted as $\sigma_{u+1}$, for decoding step $u$.
In order to build the copy target, we perform maximum matching on the transcription text $\bm{y}$ from left to right based on the dictionary $E^{b}$. 
If the token span $\bm{y}_{i,j}=(y_i, ..., y_j)$ matches the $k$-th entity $e_k$ in $E^{b}$, then we set the copy target $\sigma_{i}=e_{k}$, and $\sigma_{i+1 \sim j}=\varnothing$. This indicates that the model can copy the $k$-th entity from the dictionary at decoding step $i-1$, but cannot copy any entity from decoding step $i$ to $j-1$.
When it comes to a span of length 1, i.e., $i=j$, during the left-to-right maximum matching process, we also set $\sigma_{i}$ to $\varnothing$%
\footnote{Please note that in this paper, we primarily focus on entities with a length greater than 1, and therefore only retain such entities in our dictionary. 
}.

For example, in the instance shown in Figure \ref{fig:copyne-model}, the span ``安徽'' (An hui) matches the second entity in the dictionary, and the span ``铜陵'' (Tong ling) matches the first entity in the dictionary. This means that at steps 0 and 2, the model can choose to copy the second and first entities from the dictionary, respectively. Therefore, $\sigma_{1}=e_2$ and $\sigma_{3}=e_1$, while $\sigma_{2}=\varnothing$ and $\sigma_{4}=\varnothing$.

After constructing all the copy targets $\bm{\sigma}=(\sigma_{1}, ..., \sigma_{U})$, we can compute the copy loss as follows:
\begin{equation}\label{equ:copy-loss}
    \mathcal{L}_{copy}(\bm{\sigma}) = -\sum_{u=0}^{U-1}\log P_{c}(\sigma_{u+1}|y_{\leq u})
\end{equation}
where $P_{c}(\sigma_{u+1}|y_{\leq u})$ is the copy probability computed in Equation \ref{equ:att_prob}, meaning the probability of copying entity $\sigma_{u+1}$ at decoding step $u$. 
It is worth noting that the copy loss and the bias loss in CBA have fundamental differences. The bias loss in CBA provides information to each token, including tokens within entities, about which entity to attend to. In contrast, our copy loss solely instructs the model to copy the entity from the dictionary at the first token of the entity.

Finally, the loss in our CopyNE model is formed as follows:
\begin{equation}\label{equ:final_loss}
    \mathcal{L} = \lambda \mathcal{L}_{trans}(\bm{y}) + (1-\lambda)\mathcal{L}_{ctc}(\bm{y}) + \mathcal{L}_{copy}(\bm{\sigma})
\end{equation}

\paragraph{Dictionary Construction.}
To construct the copy target and compute the copy loss, we should first build a contextual NE dictionary for training. Provided that the entities have been labeled in the dataset, we build a NE dictionary $E^{b}$ for each data batch following previous works.

Firstly, to construct $E^{b}$, we extract all entities in the instances of this batch and add them to the dictionary. 
For instances that do not contain any entities, in order to ensure an adequate number of positive examples, we randomly select one or two substrings of length 2 or 3 from the transcription and include them in the dictionary as pseudo entities.
In order to improve the ability of copying the correct entity from a wide range of entities,  
we also extract additional negative entities from the training set.
We analyze the influence of the quantity of negative entities on the model. Due to page constraints, we have included this section in \S \ref{app:neg_ne_in_training}.

\subsection{Inference}

During inference, unlike previous token-level approaches,
our model has the flexibility to predict either a token from the vocabulary or an entity from the NE dictionary.
By copying the tokens of an entity at once, our CopyNE model can avoid errors 
that occur when predicting multiple tokens separately. 
As shown in Figure \ref{fig:copyne-model}, our CopyNE model can directly copy the two entities ``安徽'' and ``铜陵'' from the dictionary.

Specifically, at step $u$, our prediction is based on both the model's probability for a token $v$, i.e., $P(v|\hat{y}{\leq u}, E)$, and the copy probability for an entity $e$, i.e., $P_{c}(e|\hat{y}_{\leq u})$. The former represents the probability of predicting a token $v$ from the token vocabulary, while the latter is normalized on all entities, originally indicating the attention probability over entity $e$, which can be naturally interpreted as the probability of copying entity $e$ from the dictionary.
To consider both probabilities on the same scale, we devise an elegant decoding strategy by 
taking use of the copy probability of $\varnothing$, i.e., $P_{c}(\varnothing|\hat{y}_{\leq u})$, and re-normalize the probabilities to create an unified searching space $Q$.
\begin{equation}\label{equ:q}
    Q(i|\hat{y}_{\leq u}) = \begin{cases} P_{c}(\varnothing|\hat{y}_{\leq u})P(i|\hat{y}_{\leq u}, E), ~~ i \in \mathcal{V} \\
    P_{c}(i|\hat{y}_{\leq u}), ~~~~~~~~~~~~ i \in E, i \neq \varnothing
                                    \end{cases}
\end{equation}

Here, to ensure the sum of the probabilities of all elements is 1, we use $P_{c}(\varnothing|\hat{y}_{\leq u})$ as a prior probability, representing the probability of the text to be transcribed has no relation with the entities in the dictionary and the text should be generated from the token vocabulary.
If the element is from the token vocabulary $\mathcal{V}$, we obtain the probability by multiplying the prior probability and the model's probability for the token. Otherwise, we use the copy probability directly.

However, in our experiments, we observe that the model occasionally selects irrelevant entities for copying.
To enhance the quality of copying, we introduce a confidence threshold $\gamma$ during decoding to filter out low-confidence copies.
Specifically, we set $P_{c}(i|\hat{y}_{\leq u})=0,i \in {E}, i \neq \varnothing$, and $P_{c}(\varnothing|\hat{y}_{\leq u})=1$ when 
$\max\{ P_{c}(i|\hat{y}_{\leq u}) | i \in {E}, i \neq \varnothing \}<\gamma$.
This means that if the model's maximum copy probability over real entities is less than $\gamma$, it is prevented from copying entities from the dictionary and instead generates tokens from the token vocabulary.
In section \ref{exp:gamma}, we discuss the influence of the $\gamma$ in detail.

Finally, we use beam search to select the best element at each step to form the final prediction\footnote{It has to be noted that the partially predicted $\hat{\bm{y}}$ is still encoded at token-level.}.
\begin{equation}
    \hat{\bm{y}} = \arg \max \limits_{\bm{y}} (\sum_{u}\log Q(i|y_{\leq u}))
\end{equation}

%% file: figs/joint-ctc-transformer.tex
\begin{figure*}[htbp]
    \centering
    \begin{minipage}{1\columnwidth}
    \centering
    \scalebox{0.6}{
    \begin{tikzpicture}[
        connect/.style={
                rounded corners=4pt,
                very thick,
                draw=black!80
            },
        arrow/.style={
                arrows = {-Straight Barb[length=0.8mm]},
                shorten >= 2pt,
                shorten <= 1.5pt,
                very thick
            },
        inner arrow/.style={
                arrows = {-Straight Barb[length=0.4mm]},
                shorten >= 2pt,
                shorten <= 2pt,
                thin,
                draw=black!50
            },
        input/.style={
                rectangle,
                rounded corners=1mm,
                thin,
                dashed,
                draw=none,
                minimum width=3.5cm,
                minimum height=0.6cm,
            },
        share/.style={
                minimum height=0.5cm,
                  fill=orange!10,
                draw={rgb,255:red,225; green,209; blue,183},
                rounded corners=2mm,
            },
        chart/.style={
                circle,
                minimum size=3mm,
                draw=white,
                fill={rgb,255:red,189; green,215; blue,237},
                inner sep=0pt
        },
        chart1/.style={
                circle,
                minimum size=3mm,
                draw=white,
                fill={rgb,255:red,30; green,78; blue,120},
                inner sep=0pt
        },
        chart2/.style={
                circle,
                minimum size=3mm,
                draw=white,
                fill={rgb,255:red,205; green,223; blue,230},
                inner sep=0pt
        },
        chart3/.style={
                circle,
                minimum size=3mm,
                draw=white,
                fill={rgb,255:red,68; green,113; blue,199},
                inner sep=0pt
        },
        vilayer/.style={
                trapezium, 
                draw, 
                trapezium left angle=60, 
                trapezium right angle=120,
                rounded corners=1mm,
        },
        task4/.style={
                minimum height=0.5cm,
                draw=black,
                rounded corners=2mm,
                fill={rgb,255:red,240; green,249; blue,244},
            },
        task3/.style={
                minimum height=0.5cm,
                draw=white,
                rounded corners=2mm,
            },
        task2/.style={
                minimum height=0.5cm,
                fill={rgb,255:red,204; green,223; blue,230},
                draw={rgb,255:red,151; green,181; blue,191},
                rounded corners=2mm,
            },
        label/.style={
                inner sep=0.5mm,
                fill=white,
                minimum height=0.5cm,
            },
        task1/.style={
                minimum height=0.5cm,
                fill={rgb,255:red,253; green,210; blue,191},
                draw={rgb,255:red,233; green,133; blue,128},
                rounded corners=2mm
            },
        inner lstm/.style={
                fill=white,
                rectangle,
                rounded corners=1mm,
                semithick,
                draw=black!50,
                fill opacity=0.8
            },
        cell/.style={
                inner sep=2mm,
                rectangle,
                rounded corners=1mm,
                semithick,
                draw=black!50,
            },
        ocell/.style={
                solid,
                minimum height=0.5cm,
                rectangle,
                rounded corners=1mm,
                thick,
            },
        dep arrow/.style={
        arrows = {-Latex[round,open,length=8pt,width=6pt]},
        shorten >= 2pt,
        shorten <= 1.5pt,
        thick
        }
        ]
        \centering
        \node[inner sep=0pt] (waveform) at (0,0)
            {\includegraphics[width=0.4\textwidth]{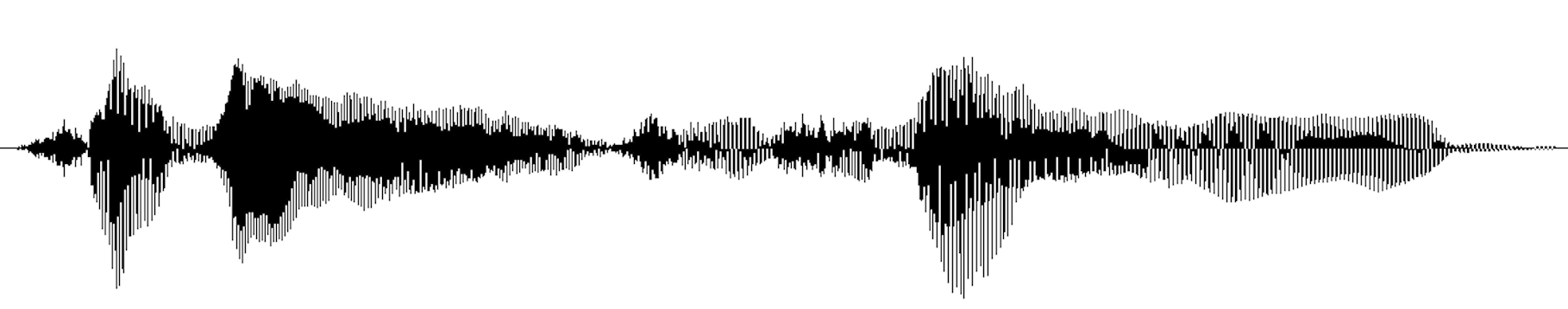}};
        \node[anchor=base] at ($(waveform.north) + (-0.3cm, 0.01cm)$) (input_x) {$\bm{X}$};
        \node [task2, ocell, anchor=south, fill opacity=0.85, minimum height=1.29cm, minimum width=2.88cm, align=center] (encoder) at ($(waveform.north) + (0, 0.8cm)$) {Audio \\ Encoder};
        \draw [arrow, connect] ($(waveform.north) + (0cm, -0.2cm)$) -- ($(encoder.south) + (0cm, 0)$);
        \node[anchor=base] at ($(encoder.north) + (-0.3cm, 0.2cm)$) (encoder_h_1) {$\bm{H}$};
        \node [task2, ocell, anchor=west, fill opacity=0.85, minimum height=1.29cm, minimum width=4.7cm, align=center] (decoder) at ($(encoder.east) + (1.6cm, 0.0cm)$) {Decoder};
        \draw [arrow, connect] ($(encoder.east) + (0.cm, 0cm)$) -- ($(decoder.west) + (0cm, 0)$);
        \node[anchor=base] at ($(encoder.east) + (+0.75cm, 0.1cm)$) (encoder_h_2) {$\bm{H}$};
        \node [share, ocell] [minimum width=2.88cm, minimum height=0.675cm, anchor=south, align=center] (ctc-layer) at ($(encoder.north) + (0, 0.7cm)$) {CTC Layer};
        \draw [arrow, connect] ($(encoder.north) + (0.cm, 0cm)$) -- ($(ctc-layer.south) + (0cm, 0)$);
        \node[anchor=base] at ($(ctc-layer.north) + (0cm, 0.7cm)$) (ctc-loss) {$\mathcal{L}_{ctc}$};
        \draw [arrow, connect] ($(ctc-layer.north) + (0.cm, 0cm)$) -- ($(ctc-loss.south) + (0cm, +0.15cm)$);

        \node[anchor=base] at ($(decoder.south west) + (+0.4cm, -0.85cm)$) (start) {<s>$_0$};
        \node[anchor=east] at ($(start.west) + (-0.01cm, 0cm)$) (decoder-input-1) {\small decoder};
        \node[anchor=base] at ($(decoder-input-1.south) + (-0.0cm, -0.2cm)$) (decoder-input-2) {\small input:};
        
        \node[anchor=west] at ($(start.east) + (+0.08cm, 0)$) (an) {安$_1$};
        \node[anchor=base] at ($(an.south) + (+0.0cm, -0.2cm)$) (AN) {\pinyin{an1}};
        
        \node[anchor=west] at ($(an.east) + (+0.12cm, 0)$) (hui) {徽$_2$};
        \node[anchor=base] at ($(hui.south) + (+0.0cm, -0.2cm)$) (HUI) {\pinyin{hui1}};
        
        \node[anchor=west] at ($(hui.east) + (+0.12cm, 0)$) (tong) {铜$_3$};
        \node[anchor=base] at ($(tong.south) + (+0.0cm, -0.2cm)$) (TONG) {\pinyin{tong2}};
        
        \node[anchor=west] at ($(tong.east) + (+0.12cm, 0)$) (ling) {陵$_4$};
        \node[anchor=base] at ($(ling.south) + (+0.0cm, -0.2cm)$) (LING) {\pinyin{ling2}};
        
        \node[anchor=west] at ($(ling.east) + (+0.1cm, 0)$) (dot) {$\ldots$};

        \draw [arrow, connect] ($(start.north) + (0.cm, -0.05cm)$) -- ($(start.south) + (0cm, +1.05cm)$);
        \draw [arrow, connect] ($(an.north) + (0.cm, -0.1cm)$) -- ($(an.south) + (0cm, +1.1cm)$);
        \draw [arrow, connect] ($(hui.north) + (0.cm, -0.1cm)$) -- ($(hui.south) + (0cm, +1.1cm)$);
        \draw [arrow, connect] ($(tong.north) + (0.cm, -0.1cm)$) -- ($(tong.south) + (0cm, +1.1cm)$);
        \draw [arrow, connect] ($(ling.north) + (0.cm, -0.1cm)$) -- ($(ling.south) + (0cm, +1.1cm)$);

        \node[anchor=base] at ($(decoder.north west) + (+0.4cm, +0.7cm)$) (tgt-an) {安$_1$};
        \node[anchor=west] at ($(tgt-an.east) + (+0.12cm, 0)$) (tgt-hui) {徽$_2$};
        \node[anchor=west] at ($(tgt-hui.east) + (+0.12cm, 0)$) (tgt-tong) {铜$_3$};
        \node[anchor=west] at ($(tgt-tong.east) + (+0.12cm, 0)$) (tgt-ling) {陵$_4$};
        \node[anchor=west] at ($(tgt-ling.east) + (+0.1cm, 0)$) (tgt-dot) {$\ldots$};

        \draw [arrow, connect] ($(tgt-an.south) + (0.cm, -0.55cm)$) -- ($(tgt-an.south) + (0cm, +0.1cm)$);
        \draw [arrow, connect] ($(tgt-hui.south) + (0.cm, -0.55cm)$) -- ($(tgt-hui.south) + (0cm, +0.1cm)$);
        \draw [arrow, connect] ($(tgt-tong.south) + (0.cm, -0.55cm)$) -- ($(tgt-tong.south) + (0cm, +0.1cm)$);
        \draw [arrow, connect] ($(tgt-ling.south) + (0.cm, -0.55cm)$) -- ($(tgt-ling.south) + (0cm, +0.1cm)$);
        \draw [arrow, connect] ($(tgt-dot.south) + (0.cm, -0.7cm)$) -- ($(tgt-dot.south) + (0cm, -0.05cm)$);

        \node [share, fill=none, draw=gray, densely dashed] [minimum width=4.5cm, minimum height=0.6cm, anchor=south] (dashed-tgt-rec) at ($(decoder.north) + (-0.25cm, 0.6cm)$) {};
        \node[anchor=east] at ($(dashed-tgt-rec.west) + (-0.0cm, 0.0cm)$) (y) {$\bm{y}$:};
        \node[anchor=base] at ($(dashed-tgt-rec.north) + (0cm, 0.7cm)$) (trans-loss) {$\mathcal{L}_{trans}$};
        \draw [arrow, connect] ($(dashed-tgt-rec.north) + (0.cm, 0cm)$) -- ($(trans-loss.south) + (0cm, +0.15cm)$);
        
    \end{tikzpicture}}
    \caption{The CTC-Transformer model.}
    \label{fig:joint-ctc-transformer-model}
    \end{minipage}
    \begin{minipage}{1\columnwidth}
    \centering
    \scalebox{0.6}{
    \begin{tikzpicture}[
        connect/.style={
                rounded corners=4pt,
                very thick,
                draw=black!80
            },
        arrow/.style={
                arrows = {-Straight Barb[length=0.8mm]},
                shorten >= 2pt,
                shorten <= 1.5pt,
                very thick
            },
        inner arrow/.style={
                arrows = {-Straight Barb[length=0.4mm]},
                shorten >= 2pt,
                shorten <= 2pt,
                thin,
                draw=black!50
            },
        input/.style={
                rectangle,
                rounded corners=1mm,
                thin,
                dashed,
                draw=none,
                minimum width=3.5cm,
                minimum height=0.6cm,
            },
        share/.style={
                minimum height=0.5cm,
                  fill=orange!10,
                draw={rgb,255:red,225; green,209; blue,183},
                rounded corners=2mm,
            },
        chart/.style={
                circle,
                minimum size=3mm,
                draw=white,
                fill={rgb,255:red,189; green,215; blue,237},
                inner sep=0pt
        },
        chart1/.style={
                circle,
                minimum size=3mm,
                draw=white,
                fill={rgb,255:red,30; green,78; blue,120},
                inner sep=0pt
        },
        chart2/.style={
                circle,
                minimum size=3mm,
                draw=white,
                fill={rgb,255:red,205; green,223; blue,230},
                inner sep=0pt
        },
        chart3/.style={
                circle,
                minimum size=3mm,
                draw=white,
                fill={rgb,255:red,68; green,113; blue,199},
                inner sep=0pt
        },
        vilayer/.style={
                trapezium, 
                draw, 
                trapezium left angle=60, 
                trapezium right angle=120,
                rounded corners=1mm,
        },
        task4/.style={
                minimum height=0.5cm,
                draw=black,
                rounded corners=2mm,
                fill={rgb,255:red,240; green,249; blue,244},
            },
        task3/.style={
                minimum height=0.5cm,
                draw=white,
                rounded corners=2mm,
            },
        task2/.style={
                minimum height=0.5cm,
                fill={rgb,255:red,204; green,223; blue,230},
                draw={rgb,255:red,151; green,181; blue,191},
                rounded corners=2mm,
            },
        label/.style={
                inner sep=0.5mm,
                fill=white,
                minimum height=0.5cm,
            },
        task1/.style={
                minimum height=0.5cm,
                fill={rgb,255:red,253; green,210; blue,191},
                draw={rgb,255:red,233; green,133; blue,128},
                rounded corners=2mm
            },
        task5/.style={
                minimum height=0.5cm,
                fill={rgb,255:red,187; green,214; blue,184},
                draw={rgb,255:red,148; green,175; blue,159},
                rounded corners=2mm
            },
        task6/.style={
                minimum height=0.5cm,
                fill={rgb,255:red,255; green,255; blue,255},
                draw={rgb,255:red,0; green,0; blue,0},
                rounded corners=2mm
            },
        inner lstm/.style={
                fill=white,
                rectangle,
                rounded corners=1mm,
                semithick,
                draw=black!50,
                fill opacity=0.8
            },
        cell/.style={
                inner sep=2mm,
                rectangle,
                rounded corners=1mm,
                semithick,
                draw=black!50,
            },
        ocell/.style={
                solid,
                minimum height=0.5cm,
                rectangle,
                rounded corners=1mm,
                thick,
            },
         hatcell/.style={
                solid,
                minimum height=0.5cm,
                rectangle,
                rounded corners=0mm,
                thick,
            },
        dep arrow/.style={
        arrows = {-Latex[round,open,length=8pt,width=6pt]},
        shorten >= 2pt,
        shorten <= 1.5pt,
        thick
        }
        ]
        \centering
        \node[inner sep=0pt] (waveform) at (0,0)
            {\includegraphics[width=0.4\textwidth]{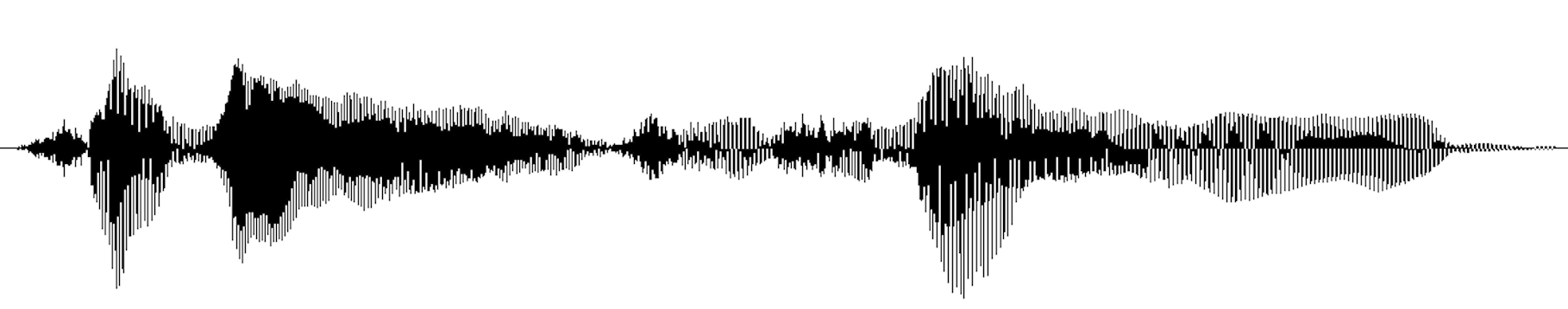}};
        \node[anchor=base] at ($(waveform.north) + (-0.3cm, 0.01cm)$) (input_x) {$\bm{X}$};
        \node [task2, ocell, anchor=south, fill opacity=0.85, minimum height=1.29cm, minimum width=2.88cm, align=center] (encoder) at ($(waveform.north) + (0, 0.8cm)$) {Audio \\ Encoder};
        \draw [arrow, connect] ($(waveform.north) + (0cm, -0.2cm)$) -- ($(encoder.south) + (0cm, 0)$);
        \node[anchor=base] at ($(encoder.north) + (-0.3cm, 0.2cm)$) (encoder_h_1) {$\bm{H}$};
        \node [task2, ocell, anchor=west, fill opacity=0.85, minimum height=1.29cm, minimum width=4.7cm, align=center] (decoder) at ($(encoder.east) + (1.6cm, 0.0cm)$) {Decoder};
        \node[anchor=base] at ($(decoder.north) + (-0.3cm, 0.2cm)$) (decoder_out) {$\bm{d}_{u}$};
        
        \draw [arrow, connect] ($(encoder.east) + (0.cm, 0cm)$) -- ($(decoder.west) + (0cm, 0)$);
        \node[anchor=base] at ($(encoder.east) + (+0.75cm, 0.1cm)$) (encoder_h_2) {$\bm{H}$};
        \node [share, ocell] [minimum width=2.88cm, minimum height=0.675cm, anchor=south, align=center] (ctc-layer) at ($(encoder.north) + (0, 0.7cm)$) {CTC Layer};
        \draw [arrow, connect] ($(encoder.north) + (0.cm, 0cm)$) -- ($(ctc-layer.south) + (0cm, 0)$);
        \node[anchor=base] at ($(ctc-layer.north) + (0cm, 0.7cm)$) (ctc-loss) {$\mathcal{L}_{ctc}$};
        \draw [arrow, connect] ($(ctc-layer.north) + (0.cm, 0cm)$) -- ($(ctc-loss.south) + (0cm, +0.15cm)$);

        \node[anchor=base] at ($(decoder.south west) + (+0.4cm, -0.85cm)$) (start) {<s>$_0$};
        \node[anchor=east] at ($(start.west) + (-0.01cm, 0cm)$) (decoder-input-1) {\small decoder};
        \node[anchor=base] at ($(decoder-input-1.south) + (-0.0cm, -0.2cm)$) (decoder-input-2) {\small input:};
        
        \node[anchor=west] at ($(start.east) + (+0.08cm, 0)$) (an) {安$_1$};
        \node[anchor=base] at ($(an.south) + (+0.0cm, -0.2cm)$) (AN) {\pinyin{an1}};
        
        \node[anchor=west] at ($(an.east) + (+0.12cm, 0)$) (hui) {徽$_2$};
        \node[anchor=base] at ($(hui.south) + (+0.0cm, -0.22cm)$) (HUI) {\pinyin{hui1}};
        
        \node[anchor=west] at ($(hui.east) + (+0.12cm, 0)$) (tong) {铜$_3$};
        \node[anchor=base] at ($(tong.south) + (+0.0cm, -0.2cm)$) (TONG) {\pinyin{tong2}};
        
        \node[anchor=west] at ($(tong.east) + (+0.12cm, 0)$) (ling) {陵$_4$};
        \node[anchor=base] at ($(ling.south) + (+0.0cm, -0.2cm)$) (LING) {\pinyin{ling2}};
        
        \node[anchor=west] at ($(ling.east) + (+0.1cm, 0)$) (dot) {$\ldots$};

        \draw [arrow, connect] ($(start.north) + (0.cm, -0.05cm)$) -- ($(start.south) + (0cm, +1.05cm)$);
        \draw [arrow, connect] ($(an.north) + (0.cm, -0.1cm)$) -- ($(an.south) + (0cm, +1.1cm)$);
        \draw [arrow, connect] ($(hui.north) + (0.cm, -0.1cm)$) -- ($(hui.south) + (0cm, +1.1cm)$);
        \draw [arrow, connect] ($(tong.north) + (0.cm, -0.1cm)$) -- ($(tong.south) + (0cm, +1.1cm)$);
        \draw [arrow, connect] ($(ling.north) + (0.cm, -0.1cm)$) -- ($(ling.south) + (0cm, +1.1cm)$);

        \node [task1, ocell] [minimum width=4.7cm, minimum height=0.6cm, anchor=south, fill opacity=0.85] (dot-att) at ($(decoder.north) + (0.0cm, 0.8cm)$) {Dot-product Attention};
        \draw [arrow, connect] ($(decoder.north) + (0.cm, 0.0cm)$) -- ($(dot-att.south) + (0cm, +0cm)$);

        \node [task5, ocell] [minimum width=2.7cm, minimum height=0.8cm, anchor=west, fill opacity=0.85, align=center] (concoder) at ($(dot-att.east) + (1.cm, 0.0cm)$) {NE \\ Encoder};
        \node[anchor=base] at ($(concoder.west) + (-0.5cm, -0.45cm)$) (concoder_out) {$\bm{Z}$};

        \node [task6, hatcell] [minimum width=2.7cm, minimum height=0.2cm, anchor=west, fill opacity=0.85, align=center] (z1) at ($(decoder.north east) + (1.cm, -0.35cm)$) {$e_0=\varnothing$};
        \node [task6, hatcell] [minimum width=2.7cm, minimum height=0.2cm, anchor=north, fill opacity=0.85, align=center] (z2) at ($(z1.south) + (0.cm, -0.01cm)$) {$e_1=$\textcolor{blue}{铜陵}};
        \node [task6, hatcell] [minimum width=2.7cm, minimum height=0.2cm, anchor=north, fill opacity=0.85, align=center] (z3) at ($(z2.south) + (0.cm, -0.01cm)$) {$e_2=$\textcolor{red}{安徽}};
        \node[anchor=base] at ($(z3.south) + (+0.cm, -0.5cm)$) (input_dictionary) {{NE dictionary}};

        \draw [arrow, connect] ($(z1.north) + (0.cm, -0.0cm)$) -- ($(concoder.south) + (0cm, +0.0cm)$);
        \draw [arrow, connect] ($(concoder.west) + (0.cm, -0.0cm)$) -- ($(dot-att.east) + (0cm, +0.0cm)$);

        \node[anchor=base] at ($(dot-att.north west) + (+0.4cm, +0.8cm)$) (tgt-an) {\textcolor{red}{安}$_1$};
        \node[anchor=west] at ($(tgt-an.east) + (+0.12cm, 0)$) (tgt-hui) {\textcolor{red}{徽}$_2$};
        \node[anchor=west] at ($(tgt-hui.east) + (+0.12cm, 0)$) (tgt-tong) {\textcolor{blue}{铜}$_3$};
        \node[anchor=west] at ($(tgt-tong.east) + (+0.12cm, 0)$) (tgt-ling) {\textcolor{blue}{陵}$_4$};
        \node[anchor=west] at ($(tgt-ling.east) + (+0.1cm, 0)$) (tgt-dot) {$\ldots$};
        \draw [arrow, connect] ($(tgt-an.south) + (0.cm, -0.55cm)$) -- ($(tgt-an.south) + (0cm, +0.08cm)$);
        \draw [arrow, connect] ($(tgt-hui.south) + (0.cm, -0.55cm)$) -- ($(tgt-hui.south) + (0cm, +0.08cm)$);
        \draw [arrow, connect] ($(tgt-tong.south) + (0.cm, -0.55cm)$) -- ($(tgt-tong.south) + (0cm, +0.08cm)$);
        \draw [arrow, connect] ($(tgt-ling.south) + (0.cm, -0.55cm)$) -- ($(tgt-ling.south) + (0cm, +0.08cm)$);
        \draw [arrow, connect] ($(tgt-dot.south) + (0.cm, -0.7cm)$) -- ($(tgt-dot.south) + (0cm, -0.08cm)$);
        \node [share, fill=none, draw=gray, densely dashed] [minimum width=4.5cm, minimum height=0.8cm, anchor=south] (dashed-tgt-rec) at ($(dot-att.north) + (-0.25cm, 0.6cm)$) {};
        \node[anchor=east] at ($(dashed-tgt-rec.west) + (-0.0cm, 0.cm)$) (y) {$\bm{y}$:};
        \node[anchor=west] at ($(dashed-tgt-rec.east) + (1.cm, 0cm)$) (trans-loss) {$\mathcal{L}_{trans}$};
        \draw [arrow, connect] ($(dashed-tgt-rec.east) + (0.cm, 0cm)$) -- ($(trans-loss.west) + (0cm, +0.cm)$);

        \node[anchor=base] at ($(tgt-an.north) + (+0.0cm, +0.7cm)$) (cp-tgt-1) {\textcolor{red}{$e_2$}};
        \node[anchor=base] at ($(tgt-hui.north) + (+0.0cm, +0.7cm)$) (cp-tgt-2) {$e_0$};
        \node[anchor=base] at ($(tgt-tong.north) + (+0.0cm, +0.7cm)$) (cp-tgt-3) {\textcolor{blue}{$e_1$}};
        \node[anchor=base] at ($(tgt-ling.north) + (+0.0cm, +0.7cm)$) (cp-tgt-4) {$e_0$};
        \node[anchor=base] at ($(tgt-dot.north) + (+0.0cm, +0.9cm)$) (cp-tgt-5) {$\ldots$};
        \node [share, fill=none, draw=gray, densely dashed] [minimum width=4.5cm, minimum height=0.6cm, anchor=south] (dashed-copy-rec) at ($(dashed-tgt-rec.north) + (-0.cm, 0.35cm)$) {};
        \node[anchor=east] at ($(dashed-copy-rec.west) + (-0.0cm, 0.cm)$) (cp) {$\bm{\sigma}$:};
        \node[anchor=west] at ($(dashed-copy-rec.east) + (1.cm, 0cm)$) (copy-loss) {$\mathcal{L}_{copy}$};
        \draw [arrow, connect] ($(dashed-copy-rec.east) + (0.cm, 0cm)$) -- ($(copy-loss.west) + (0cm, +0.cm)$);

    \end{tikzpicture}}
    \caption{Our CopyNE model.}
    \label{fig:copyne-model}
    \end{minipage}
\end{figure*}

%% file: content/exp.tex
\section{Experiments}
\subsection{Experimental Setup}
\paragraph{Datasets.}
Experiments on Chinese Mandarin are conducted on two widely used datasets, Aishell \cite{bu-2017-aishell} and ST-cmds\footnote{https://www.openslr.org/38/}. 
We use the Eng dataset released by \citet{YadavG0S20} to perform experiments on English.
Furthermore, to compare the performance of different methods in contextual ASR scenarios where speeches contain entities, we extract instances containing entities from the dev and test sets, forming the corresponding ``$*$-NE'' datasets.
Detailed introduction about the datasets can be found in \S \ref{app:data}.

\paragraph{NE Dictionary.}
Aishell and ST-cmds were released without entity annotations. In contrast, the Eng dataset was simultaneously released with audio, transcribed text, and corresponding entity annotations.
\citet{chen-2022-aishell} further annotated entities for Aishell. 
So, in our experiments with Aishell and Eng, we use the releated 
entities to build the NE dictionary. 
For ST-cmds, we use HanLP\footnote{https://github.com/hankcs/HanLP} to get three types of entities: person, location, and organization.

\paragraph{Evaluation Metrics.}
Character error rate (CER) and word error rate (WER) are used to assess the overall performance of models in Mandarin and English ASR tasks.
In this paper, to evaluate the model's entity transcription accuracy, we also employ NE-CER and NE-WER metrics \cite{han-2021-cif}. We align the predicted hypothesis and reference using the minimum edit distance algorithm, and subsequently calculate NE-C(W)ER by measuring the C(W)ER between the entity text in the reference and its counterpart in the hypothesis.

\input{tables/overall_res}
\paragraph{Parameter Setting.}
The parameter setting in our work is the same as that in most previous works, and the detailed descriptions can be found in \S \ref{app:para}.
To ensure a fair comparison with prior works, we carefully reproduced  the CLAS \cite{pundak-2018-clas} and CBA \cite{zhang-2022-cba}.
Moreover, to verify the effectiveness of our approach on pre-trained large models, we also conducted experiments on  OpenAI  Whisper \cite{radford-2022-whisper}.
Specifically, we use the Whisper model as our transformer encoder and decoder. 
We choose seeds randomly to run models for 3 times and report the average results.

\subsection{Results} \label{exp:gamma}
\input{figs/cp_threshold}

\paragraph{Analysis about $\gamma$.}
We first investigate the influence of the copy threshold $\gamma$ during inference.
Figure \ref{fig:cp_threshold} illustrates how the CER changed on the Aishell dev and Aishell-NE dev with different $\gamma$ values.
Our findings reveal that when the threshold is low, the CER is high, indicating that copying results in more errors  when the model copies entities with low confidence.
As we increase the $\gamma$, the CER decreases, indicating improved reliability of our CopyNE when the model had higher confidence.
However, when the threshold becomes too high (above 0.9), the model has fewer opportunities to choose to copy entities, resulting in a higher CER.
This happens because it becomes more difficult for the model to trigger the copy mechanism.
So, we set $\gamma$ to 0.9 for all experiments and discussions to enhance the robustness of our model.

\input{tables/ne-cer}

\paragraph{Results on Chinese.}
Table \ref{table:overall-cer} and \ref{table:ne-cer} show the CER and NE-CER of different models on the Chinese dataset. In Table \ref{table:overall-cer}, we note that while our primary focus is improving transcription of NEs, we also achieve significant improvements in overall text transcription. 
Without Whisper, our CopyNE model outperforms the previous CBA approach with a 3.2\% relative CER reduction on the Aishell Test and 7.7\% on the ST-cmds Test. 
In contextual ASR scenarios, the improvements are even more pronounced, with a 13.5\% relative CER reduction on the Aishell-NE Test and 20.8\% on the ST-cmds-NE Test. 
Even with the powerful Whisper, our CopyNE consistently excels, especially on the ST-cmds dataset, with relative reductions of 8.6\% and 15.7\% on the two test sets, respectively. 
Additionally, we observed that CLAS performs well on Aishell, closely matching CopyNE, but its performance on ST-cmds is comparatively weaker, sometimes even worse than the Whisper baseline, a reverse pattern also seen with CBA. In contrast, CopyNE consistently performs well across different datasets, demonstrating its better adaptability.

We also present an improved model, i.e. CopyNE$^{\dagger}$, which features a more powerful conformer encoder. The results show that CopyNE$^{\dagger}$ can achieve further improvements compared to CopyNE.

In this paper, our main goal is improving the transcription of NEs.
From the results presented in Table \ref{table:ne-cer}, it is evident that our approach exhibits significant improvements in entity transcription compared to previous methods. 
When not using Whisper, our CopyNE model achieves an impressive relative NE-CER reduction of 55.4\% on the Aishell-NE Test and 53.9\% on the ST-cmds-NE Test.
Even based on the powerful Whisper model, our CopyNE continues to achieve remarkable improvements, with a relative NE-CER reduction of 25.4\% and 26.7\% on the two test sets. 
This demonstrates that copying entities from the dictionary significantly improves the accuracy of transcribing entities.

\input{tables/eng-res}
\paragraph{Results on English.}
Whisper \cite{radford-2022-whisper} has shown strong performance in English, so we directly use it as our baseline for experiments on English. As seen in Table \ref{table:eng-res}, CopyNE still outperforms other methods, achieving a 5.2\% relative WER reduction compared to CLAS in the general scenario on the Eng test dataset. In contextual scenarios, CopyNE demonstrates 6.4\% relative WER reductions and 16.8\% relative NE-WER reductions. Additionally, we observed that CBA lags behind the Whisper baseline.
We suspect that this might be due to its approach of encouraging the model to generate entity tokens by modifying Whisper's output logits, which can disrupt the model's overall probability distribution, especially given Whisper's strong fit on English data.
On the contrary, our CopyNE is more stable.

\subsection{Impact of the NE dictionary} \label{exp:noisy}
Following previous works \cite{pundak-2018-clas, han-2021-cif}, we report the main results using exact NE dictionaries from the test sets.
However, when collecting dictionaries in real scenarios, to ensure the coverage, many unrelated noisy NEs are inevitably added to the dictionary. To analyze the impact of noisy entities on CopyNE, we extract entities from the training set that are not included in the test set as noisy NEs.
From the corresponding $\times2$, $\times3$, and $\times4$ rows in Table \ref{table:add_noisy_nes}, we can see that the introduction of noisy NEs results in a reduction in the model's performance. Nevertheless, even with the addition of 6k noisy NEs, resulting in the dictionary size being quadrupled ($\times4$), CopyNE continues to outperform CLAS and CBA, despite their reliance on the precise dictionary.

In the more rare cases where some NEs are out of the dictionary (OOD), to analyze CopyNE's performance in OOD scenarios, we designate some low-frequency NEs from the original dictionary as OOD NEs. These NEs are removed, and decoding is performed using the remaining NEs. From the relevant rows in Table \ref{table:add_noisy_nes}, it can be observed that this primarily impacts NE-CER since CopyNE cannot copy missing NEs. However, even when the OOD proportion reaches 15\% (Dict size = $\times$0.85), CopyNE still shows commendable performance.

\input{tables/add_noise_nes}

\subsection{Qualitative Analysis}
\input{tables/less_cases}

CopyNE demonstrates significant improvements. To gain further insight into CopyNE's performance, we conduct a qualitative analysis of its generations.
Table \ref{table:cases} shows examples of transcriptions from different ASR models. 
We can see that in the second example, where CBA successfully identified the correct person entity ``杨丙卿'' from the dictionary and produced a transcription that is close to gold, it still made a mistake by transcribing ``炳'' instead of ``丙''  due to the same pronunciation (\pinyin{bing3}). 
In contrast, CopyNE can copy all the tokens of the entity from the NE dictioanry.
For example, in the third example ``冈山的桃太郎体育馆'' (The Taotailang gym in Gangshan), 
CopyNE directly copies the location entity ``冈山'' (Gangshan) and the organization entity ``桃太郎体育馆'' (Taotailang gym), achieving a completely correct transcription.

%% file: tables/overall_res.tex
\begin{table*}[!tb]
    \centering
    \setlength{\tabcolsep}{6.5pt}
    \begin{tabular}{l rrrr rrrr}
         \toprule
        \multirow{2}{*}{Model} &
        \multicolumn{2}{c}{Aishell} & \multicolumn{2}{c}{Aishell-NE} & \multicolumn{2}{c}{ST-cmds} &  \multicolumn{2}{c}{ST-cmds-NE}\\
        \cmidrule(lr){2-3} \cmidrule(lr){4-5} \cmidrule(lr){6-7} \cmidrule(lr){8-9}
        & Dev & Test & Dev & Test & Dev & Test & Dev & Test \\
        \hline
        Joint CTC-Transformer & \textcolor{white}{0}6.12 & \textcolor{white}{0}6.70 & \textcolor{white}{0}7.36 & \textcolor{white}{0}9.00 & 10.63 & 10.56 & 13.67 & 13.63 \\
        CLAS \cite{pundak-2018-clas} & \textcolor{white}{0}6.04 & \textcolor{white}{0}6.72 & \textcolor{white}{0}7.06 & \textcolor{white}{0}8.73 & 10.10 & 10.09 & 12.64 & 12.85 \\
        CBA \cite{zhang-2022-cba} & \textcolor{white}{0}6.11 & \textcolor{white}{0}6.56 & \textcolor{white}{0}6.73 & \textcolor{white}{0}8.00 & 10.73 & 10.72 & 12.69 & 12.43 \\
        CopyNE & \textcolor{white}{0}5.59 & \textcolor{white}{0}6.35 & \textcolor{white}{0}5.36 & \textcolor{white}{0}6.92 & \textcolor{white}{0}9.76 & \textcolor{white}{0}9.89 & \textcolor{white}{0}9.90 & \textcolor{white}{0}9.84 \\
        CopyNE$^{\dagger}$ & \textcolor{white}{0}\bf4.49 & \textcolor{white}{0}\bf5.03 & \textcolor{white}{0}\bf4.31 & \textcolor{white}{0}\bf5.05 & \textcolor{white}{0}\bf7.78 & \textcolor{white}{0}\bf7.80 & \textcolor{white}{0}\bf7.71 & \textcolor{white}{0}\bf7.40 \\
        
        \hline
        Whisper & 5.28 & 5.97 & 6.32 & 7.68 & 9.14 & 9.06 & 12.22 & 12.35 \\
        \textcolor{white}{xxx}+ CLAS & \bf4.50 & 5.23 & \bf5.30 & 6.72 & 9.10 & 9.20 & 12.12 & 12.25 \\
        \textcolor{white}{xxx}+ CBA & 5.41 & 6.06 & 6.13 & 7.60 & 7.96 & 7.87 & 10.03 & 9.96 \\
        \textcolor{white}{xxx}+ CopyNE & 4.71 & \bf5.10 & 5.40 & \bf6.42 & \bf7.35 & \bf7.19 & \bf8.98 & \bf8.40 \\
        \bottomrule
    \end{tabular}
    \caption{CER on Chinese datasets in general scenarios (Aishell, ST-cmds) and contextual scenarios (Aishell-NE, ST-cmds-NE). $^{\dagger}$ means the model with an improved 12-layer Conformer \cite{gulati-2020-conformer} encoder and averages the parameters of the best 10 epochs when decoding.} 
    \label{table:overall-cer}
\end{table*}

%% file: figs/cp_threshold.tex
\begin{figure}[!tb]
    \centering
    \scalebox{0.6}{
    \begin{tikzpicture}
        \centering
        \begin{axis}[
            xlabel={$\gamma$},
            ylabel={CER(\%)},
            xmin=0,xmax=1,
            ymin=5,ymax=7.8,
            xtick={0, 0.1, 0.2, 0.3, 0.4, 0.5, 0.6, 0.7, 0.8, 0.9, 1},
            ytick={5, 5.5, 6, 6.5, 7, 7.5},
            ymajorgrids=true,
            major grid style={line width=.2pt,draw=gray!50,densely dashed},
            minor grid style={line width=.1pt,line length=.02pt draw=gray!30},
            legend pos=north east,
            ]
            \addplot[color=blue, line width=1.2pt, mark=diamond] coordinates {
            (0,7.05)
            (0.1, 7.03)
            (0.2, 6.99)
            (0.3, 6.90)
            (0.4, 6.68)
            (0.5, 6.40)
            (0.6, 5.99)
            (0.7, 5.67)
            (0.8, 5.50)
            (0.9, 5.36)
            (1, 6.40)
            };
            \node [draw=white, fill=none, anchor=east] at (axis cs:1,6.4) {6.40};
            \addlegendentry{Aishell-NE}
            \addplot[color=green, line width=1.2pt, mark=triangle] coordinates {
            (0,7.66)
            (0.1, 7.58)
            (0.2, 7.45)
            (0.3, 7.29)
            (0.4, 7.04)
            (0.5, 6.69)
            (0.6, 6.34)
            (0.7, 6.03)
            (0.8, 5.78)
            (0.9, 5.59)
            (1, 5.76)
            };
            \node [draw=white, fill=none, anchor=east] at (axis cs:1,5.76) {5.76};
            \addlegendentry{Aishell}
        \end{axis}
    \end{tikzpicture}}
    \caption{Effect of the Confidence Threshold $\gamma$.}
    \label{fig:cp_threshold}
\end{figure}

%% file: tables/ne-cer.tex
\begin{table}[!tb]
\begin{small}
    \centering
    \setlength{\tabcolsep}{3pt}
    \begin{tabular}{l cc cc}
         \toprule
        \multirow{2}{*}{Model} & \multicolumn{2}{c}{Aishell-NE} & \multicolumn{2}{c}{ST-cmds-NE}\\
        \cmidrule(lr){2-3} \cmidrule(lr){4-5}
        & Dev & Test & Dev & Test \\
        \hline
        Joint CTC-Transformer & 11.64 & 14.03 & 21.63 & 21.41 \\
        CLAS \cite{pundak-2018-clas} & 11.24 & 13.12 & 19.70 & 20.10 \\
        CBA \cite{zhang-2022-cba} & \textcolor{white}{0}7.78 & \textcolor{white}{0}9.44 & 15.72 & 15.92 \\
        CopyNE & \textcolor{white}{0}\bf3.00 & \textcolor{white}{0}\bf4.21 & \textcolor{white}{0}\bf7.60 & \textcolor{white}{0}\bf7.34 \\
        \textcolor{white}{x} \emph{w/o} Dict repr $\bm{r}_{u}$ & \textcolor{white}{0}4.05 & \textcolor{white}{0}5.27 & \textcolor{white}{0}9.35 & \textcolor{white}{0}9.21 \\
        \hline
         Whisper & 10.31 & 12.24 & 21.30 & 21.83 \\
         \textcolor{white}{xxx}+ CLAS &  \textcolor{white}{0}8.97 & 11.64 & 20.82 & 21.07 \\
         \textcolor{white}{xxx}+ CBA &  \textcolor{white}{0}9.13 & 11.79 & 15.91 & 15.41 \\
        \textcolor{white}{xxx}+ CopyNE &  \textcolor{white}{0}\bf6.74 & \textcolor{white}{0}\bf{8.79} & \bf11.93 & \bf11.29 \\
        \bottomrule
    \end{tabular}
    \caption{NE-CER (\%) on the Chinese datasets.} 
    \label{table:ne-cer}
\end{small}
\end{table}

%% file: tables/eng-res.tex
\begin{table}[!tb]
\begin{small}
    \centering
    \setlength{\tabcolsep}{4.pt}
    \begin{tabular}{l rr rr rr}
         \toprule
        \multirow{2}{*}{Model} &
        \multicolumn{2}{c}{Eng\tiny{$_{\mathrm{.W}}$}} & \multicolumn{2}{c}{Eng-NE\tiny{$_{\mathrm{.W}}$}} & \multicolumn{2}{c}{Eng-NE\tiny{$_{\mathrm{.NW}}$}} \\
        \cmidrule(lr){2-3} \cmidrule(lr){4-5} \cmidrule(lr){6-7} 
        & Dev & Test & Dev & Test & Dev & Test\\
        \hline
        Whisper & 8.54 & 8.73 & 8.53 & 8.71 & 28.16 & 26.61 \\
        \textcolor{white}{xx}+ CLAS & 7.90 & 8.28 & 7.86 & 8.31 & 27.23 & 26.55 \\
        \textcolor{white}{xx}+ CBA & 9.17 & 9.52 & 9.21 & 9.49 & 30.01 & 30.82 \\
        \textcolor{white}{xx}+ CopyNE & \bf7.47 & \bf7.85 & \bf7.42 & \bf7.78 & \bf23.29 & \bf22.09 \\
        \bottomrule
    \end{tabular}
    \caption{Results on the English datasets. W and NW denote WER and NE-WER respectively.} 
    \label{table:eng-res}
\end{small}
\end{table}

%% file: tables/add_noise_nes.tex
\begin{table}[!tb]
\begin{small}
    \centering
    \setlength{\tabcolsep}{2.pt}
    \begin{tabular}{l cc cc}
         \toprule
         \multirow{2}{*}{Dict size} & \multicolumn{2}{c}{Dev} & \multicolumn{2}{c}{Test}\\
        \cmidrule(lr){2-3} \cmidrule(lr){4-5}
        & CER & NE-CER & CER & NE-CER \\
        \hline
        $\times$0.85 & \textcolor{white}{0}5.65 & \textcolor{white}{0}4.27 & 7.29 & \textcolor{white}{0}5.83 \\
        $\times$0.90 & \textcolor{white}{0}5.56 & \textcolor{white}{0}3.90 & 7.15 & \textcolor{white}{0}5.26 \\
        $\times$0.95 & \textcolor{white}{0}5.44 & \textcolor{white}{0}3.37 & 7.01 & \textcolor{white}{0}4.68 \\
        $\times$1 & \textcolor{white}{0}\bf5.36 & \textcolor{white}{0}\bf3.00 & \bf6.92 & \textcolor{white}{0}\bf4.21 \\
        $\times$2 & \textcolor{white}{0}5.61 & \textcolor{white}{0}3.50 & 7.12 & \textcolor{white}{0}4.82 \\
        $\times$3 & \textcolor{white}{0}5.85 & \textcolor{white}{0}3.92 & 7.39 & \textcolor{white}{0}5.18 \\
        $\times$4 & \textcolor{white}{0}6.02 & \textcolor{white}{0}4.09 & 7.66 & \textcolor{white}{0}5.56 \\
        \bottomrule
    \end{tabular}
    \caption{The impact of the NE dictionary.} 
    \label{table:add_noisy_nes}
\end{small}
\end{table}

%% file: tables/less_cases.tex
\begin{table}[tb]
    \begin{small}
    \centering
    \setlength{\tabcolsep}{2.0pt}
    \begin{tabular}{l|l|p{1.4cm}}
         \toprule
        & Transcriptions & Dictionary \\
        \cline{1-3}
        English & A company in Yangluo... & \multirow{15}{1.4cm}{
                                                \begin{enumerate}[]
                                                \item \scriptsize{阳逻 (Yangluo)}
                                                \item \scriptsize{杨丙卿 (Yang Bingqing)}
                                                \item \scriptsize{冈山 (Gangshan)}
                                                \item \scriptsize{桃太郎体育馆 (Taotailang gym)}
                                                \end{enumerate}}\\
        Gold & 阳逻的一家公司... & \\
        CLAS & \textcolor{red}{扬罗}的一家公司... & \\
        CBA & 阳\textcolor{red}{罗}的一家公司... & \\
        CopyNE & $\left[\textit{阳逻}\right]$的一家公司... & \\
        \cline{1-2}
        English & Yang Bingqing served as manager... & \\
        Gold & 杨丙卿担任经理... & \\
        CLAS & 杨\textcolor{red}{澄清}担任经理... & \\
        CBA & 杨\textcolor{red}{炳}卿担任经理... & \\
        CopyNE & $\left[\textit{杨丙卿}\right]$担任经理... & \\

        \cline{1-2}
        English & The Taotailang gym in Gangshan. & \\
        Gold & 冈山的桃太郎体育馆 & \\
        CLAS & 冈山的\textcolor{red}{淘汰}郎体育馆 & \\
        CBA & \textcolor{red}{刚}山的\textcolor{red}{淘汰狼}体育馆 & \\
        CopyNE & $\left[\textit{冈山}\right]$的$\left[\textit{桃太郎体育馆}\right]$ & \\
        
        \bottomrule
    \end{tabular}
    \caption{Generations of different models. Red text indicates errors, while text enclosed in square brackets represents entities that were copied from the dictionary.} 
    \label{table:cases}
    \end{small}
\end{table}

%% file: content/related_work.tex
\section{Related Works}
\paragraph{Contextual ASR.}
Researchers have explored various approaches to help models in the contextual ASR scenario, with the primary approaches being the utilization of external dictionaries and language models (LMs).
CLAS \cite{pundak-2018-clas} was the first to introduce the use of dictionary to aid in prediction.
\citet{alon-2019-hard-example} extend CLAS by adding phonetically similar alternative terms to the dictionary as negative examples, aiming to improve the model's ability to distinguish entities with similar pronunciations. 
\citet{huber-2021-instant} propose to utilize the representation of a single entry in the dictionary that is most relevant to the current decoding status.
\citet{char-level-concder-2023} propose to apply the character-based NE encoder to better capture acoustic features useful for transcribing rare entities.
Different from our CopyNE, these methods all treat entities as individual tokens, which may result in incomplete NE transcriptions.

LMs trained on large-scale text data can learn rich linguistic and contextual knowledge, and thus can be used to assist contextual ASR.
There are typically two approaches to leverage LMs for contextual ASR. 
The first approach involves using the dedicated LM to  encourage the generation of entity tokens during decoding. \cite{Novak-2012-DynamicGW, aleksic-2015-bringing, zhao-2019-shallow}.
The second approach is multi-modal pre-training.
Researchers have explored joint pre-training of speech and text models, aiming to leverage information from both modalities, and have achieved promising results \cite{chung-etal-2021-splat, ao-etal-2022-speecht5, zhang-etal-2022-speechut}.
However, compared to using contextual dictionaries, models that rely on LMs tend to have much more parameters, which means that training and deploying require more time and computational resources.

\paragraph{The Copy Mechanism.}

The copy mechanism can be traced back to the pointer network \cite{vinyals-2015-pointer}, which can predict output sequences from the input. 
The copy mechanism \cite{gu-etal-2016-incorporating} extends the pointer network by enabling the model to generate sequences that are not present in the input.
According to the source of copying, it can be divided into copying from input text, from document, and from external dictionary.

\emph{Copying from Input Text.} The copy mechanism is commonly used to copy text from input text.
For instance, in text summarization tasks, it is common to 
employ the copy mechanism to copy keywords from the input text \cite{cheng-lapata-2016-neural, xu-etal-2020-self}.
In grammatical error correction tasks, where 
only a small portion requires correction, the copy mechanism is used to copy the correct text from the input text \cite{zhao-etal-2019-improving}.

\emph{Copying from Document.} In addition to copying from input text, the copy mechanism can be employed to copy text from other texts when the input text is not available. \citet{lan-2023-copy} introduced the copy mechanism in decoder-only language models, where text fragments are selected from a vast amount of documents to generate the target text.

\emph{Copying from External Dictionary.} 
In this paper, we introduce a systematic framework, which seamlessly integrates the process of copying from an external dictionary to aid in generation. We believe that our framework can also be applied to other generation tasks.

%% file: content/conclusion.tex
\section{Conclusion}
In this paper, we consider entities as indivisible elements and introduce a copy mechanism into ASR for the first time to assist in transcribing entities. We devise a systematic copy framework that can copy all the tokens of an entity from the NE dictionary at once, preserving the token span as a complete entity.
Our approach demonstrates substantial improvements on both English and Chinese datasets.
In summary, CopyNE represents a significant advancement in contextual ASR,  providing a promising direction in this field.

%% file: content/limit.tex
\section*{Limitations}
From our experiments, we have found that an excessive number of noisy entities can impact the performance. 
As part of our future work, we intend to explore methods for dynamically filtering out interfering entities from the dictionary during the decoding process.

%% file: content/ack.tex
\section*{Acknowledgements}
We would like to thank the anonymous reviewers
for their valuable comments.
This work was supported by National Natural Science Foundation of China (Grant No. 62176173 and 62336006), and a Project Funded by the Priority Academic Program Development of Jiangsu Higher Education Institutions.

%% file: tables/dataset.tex
\begin{table}[!htb]
\begin{small}
    \centering
    \setlength{\tabcolsep}{3.pt}
    \begin{tabular}{l ccc ccc}
         \toprule
        \multirow{2}{*}{Dataset} &
        \multicolumn{2}{c}{Train} & \multicolumn{2}{c}{Dev} & \multicolumn{2}{c}{Test} \\
        \cmidrule(lr){2-3} \cmidrule(lr){4-5} \cmidrule(lr){6-7}
        & Sent & NE & Sent & NE & Sent & NE \\
        \hline
        Aishell & 119919 & 14241 & 14326 & 2194 & 7176 & 1186 \\
        Aishell-NE & 119919 & 14241 & 4949 & 2194 & 2244 & 1186 \\
        ST-cmds & 82080 & 17376 & 10260 & 3029 & 10260 & 3124 \\
        ST-cmds-NE & 82080 & 17376 & 3285 & 3029 & 3241 & 3124 \\
        Eng & 64570 & 11858 & 3100 & 2568 & 3100 & 2508 \\
        Eng-NE & 64570 & 11858 & 2677 & 2568 & 2690 & 2508 \\
        \bottomrule
    \end{tabular}
    \caption{Statistics of the used datasets.
} 
    \label{table:dataset}
\end{small}
\end{table}

%% file: figs/beta.tex
\begin{figure}[!htb]
    \centering
    \scalebox{0.6}{
    \begin{tikzpicture}
        \centering
        \begin{axis}[
                    ybar,
                    bar width=0.4cm,
                    width=8cm,
                    height=8cm,
                    xlabel={},
                    ylabel={(\%)},
                    ylabel style={yshift=-5pt}, %
                    ymin=2.8,
                    ymax=6,
                    xtick=data,
                    xtick={0,1},
                    xticklabels={CER, NE-CER},
                    xticklabel style={align=center}, %
                    ymajorgrids=true,
                    enlarge x limits=0.5,
                    major grid style={line width=.2pt,draw=gray!50,densely dashed},
                    minor grid style={line width=.1pt,line length=.02pt draw=gray!30},
                    legend pos=north east,
                    nodes near coords, %
                    nodes near coords align={vertical},
                    nodes near coords style={font=\tiny}, %
                ]
                \addplot[fill=blue!30] coordinates {
                (0, 5.78)
                (1, 3.44)
                };
                \addplot[fill=red!30] coordinates {
                (0, 5.54)
                (1, 3.02)
                };
                \addplot[fill=green!30] coordinates {
                (0, 5.36)
                (1, 3.00)
                };
                \addplot[fill=yellow!30] coordinates {
                (0, 5.59)
                (1, 3.57)
                };
                \legend{$\beta=0$, $\beta=1$, $\beta=2$, $\beta=3$}
                \end{axis}
    \end{tikzpicture}}
    \caption{Effect of $\beta$.}
    \label{fig:beta}
\end{figure}